\documentclass[a4paper]{article}

\usepackage{INTERSPEECH2022}

\usepackage{hyperref}
\usepackage{pdfpages}

\usepackage{amsmath,graphicx,amsfonts, subcaption}
\usepackage{xcolor}
\usepackage[normalem]{ulem}

\title{Unsupervised Source Separation via Bayesian Inference\\ in the Latent Domain}

\name{
Michele Mancusi$^{*1}$\thanks{* Equal contribution}, Emilian Postolache$^{* 1}$, Giorgio Mariani$^{1}$, Marco Fumero$^{1}$, Andrea Santilli$^{1}$, Luca Cosmo$^{2,3}$, Emanuele Rodolà$^{1}$
}

\address{ $^1$Sapienza University of Rome, $^2$Ca' Foscari University of Venice, $^3$University of Lugano}
\email{mancusi@di.uniroma1.it}
\begin{document}


\maketitle

\begin{abstract}
State of the art audio source separation models rely on supervised data-driven approaches, which can be expensive in terms of labeling resources. On the other hand, approaches for training these models without any direct supervision are typically high-demanding in terms of memory and time requirements, and remain impractical to be used at inference time. We aim to tackle these limitations by proposing a simple yet effective unsupervised separation algorithm, which operates directly on a latent representation of time-domain signals. Our algorithm relies on deep Bayesian priors in the form of pre-trained autoregressive networks to model the probability distributions of each source. We leverage the low cardinality of the discrete latent space, trained with a novel loss term imposing a precise arithmetic structure on it, to perform exact Bayesian inference without relying on an approximation strategy. 
We validate our approach on the Slakh dataset \cite{manilow2019}, demonstrating results in line with state of the art supervised approaches while requiring fewer resources with respect to other unsupervised methods.
\end{abstract}


\section{Introduction}
\label{sec:introduction}
Generative models have reached promising results in a wide range of domains, including audio, and can be used to solve different tasks in unsupervised learning. A relevant problem in the musical domain is the task of source separation of different instruments. Given the sequential nature of music and the high variability of rhythm,  timbre and melody, autoregressive models \cite{larochelle11} represent a popular and effective choice  to process data on such domain, showcasing high multi-modality in the modeled probability distributions. The widely adopted WaveNet autoregressive architecture \cite{oord2016} works in the temporal domain. Given that audio signals are typically sampled at high frequencies (e.g. $44$ kHz) for music, the choice of modeling the data distribution directly in the time domain leads to short contexts being captured by neural computations and quick saturation of memory. Nevertheless, existing unsupervised approaches for source separation operate in the time domain \cite{jayaram2021}. In order to capture longer contexts and to reduce memory burden, different quantization schemes have been introduced for autoregressive models \cite{oord2017,razavi2019}, where chunks in time are mapped to sequences of latent tokens belonging to a small vocabulary. OpenAI's Jukebox \cite{dhariwal:2020} follows this approach and excels as an architecture that can capture very long contexts, generating highly consistent tracks.
Leveraging the useful properties of this architecture, we propose a novel approach to unsupervised source separation that works directly on quantized latent domains. 

Our contributions can be summarized as follows:
\begin{enumerate}
    \item We perform source separation applying exact Bayesian inference directly in the latent domain, exploiting the relative small size of the latent dictionary. We do not rely on any approximation strategy, such as variational inference or Langevin dynamics. 
    \item We introduce LQ-VAE: a quantized autoencoder trained with a novel loss that  imposes an algebraic structure on the discrete latent space. This allows us to alleviate noisy and distorted samples which arise from a vanilla quantization approach.
 \end{enumerate}

\section{Related work}
\label{sec:related}

The problem of source separation has been classically tackled in an unsupervised fashion \cite{comon:1994}, where the sources to be separated from a mixture signal are unknown \cite{Smaragdis:2014}.
With the advent of deep learning, most source separation tasks applied to musical data started relying on supervised learning, training models on data with known correspondence between sources. Recently, following the success of deep generative models, there has been a renewed interest in unsupervised methods.
\subsection{Supervised source separation}
Supervised source separation aims to map high dimensional observations of audio mixtures to a smaller dimensional space and apply, explicitly or implicitly, a mask to filter out the sources from the latent representation of the mixtures in a supervised way.
Most of these works can be divided into \emph{frequency-domain} or \emph{waveform-domain} approaches. 
The former \cite{Roweis:2000} operate on the spectral representation of the input mixtures.
This line of works has highly benefited from  the incoming of deep learning techniques from simple fully connected networks \cite{Uhlich:2015}, LSTM \cite{Uhlich:2017}, and CNN coupled with recurrent approaches \cite{liu:2018,Takashi2018}.
Recent approaches such as \cite{choi:2020} and \cite{takahashi:2020} hold the state of the art in music source separation over the dataset MUSDB18 \cite{musdb18}, by respectively extending the conditional U-net architecture of \cite{meseguer-brocal2019} to  multi-source  separation, and by exploiting multi-dilated convolution that applies different dilation factors in each layer to model different resolutions simultaneously.
In contrast, waveform domain approaches process the mixtures directly in the time domain to overcome phase estimation, which is necessary when converting the signal from the frequency domain. The method of \cite{defossez2019} performs in line with the state of the art by extending a WaveNet-like architecture, coupled with an LSTM in the latent space.

The main limitation of these state-of-the-art methods for audio source separation is that they require large amounts of fully separated, labeled data to perform the training.

\subsection{Unsupervised source separation}

Recent approaches in unsupervised source separation leverage self-supervised learning. A prominent baseline is MixIt \cite{wisdom2020}, which trains a model by trying to separate sources from a mixture of mixtures. Although promising, such model suffers from the \emph{over-separation} problem, where at test time a number of sources that is greater than those present in the mixture are estimated. As such, stems can be split across different output tracks. Generative approaches instead overcome this problem by imposing that a model should output an individual stem.

Closer to our work, \cite{narayanaswamy2020unsupervised} proposes to leverage generative priors in the form of GANs trained on individual sources. They use projected gradient descent optimization to search in the source-specific latent spaces and effectively recover the constituent sources in the time domain. Although promising, GANs suffer from modal collapse, so their performance is limited in the musical domain, where variability is abundant.
\cite{jayaram2021} proposes to use Langevin dynamics on the global log-likelihood of the audio sequences to parallelize the sampling procedure of autoregressive
models used as Bayesian priors. 
This approach produces good results but with a high computational cost due to the need of training distinct models for each noise level, and due to the costly optimization procedure in the time domain.

Differently, our inference procedure has much lower computational and memory requirements, allowing us to efficiently run the model on a single GPU. In addition, we can perform exact Bayesian inference without relying on an approximation scheme of the posterior (e.g., its score).

\section{Background}

\label{sec:background}
   In this section we briefly introduce the background concepts necessary to understand our architecture, which builds upon  \cite{dhariwal:2020}. The overall architecture can be split into two parts: (i) a quantization module mapping the input sequences to a discrete latent space, and (ii) an autoregressive prior (one per source) which models the distribution of a given source in the discrete latent space. We point the reader to \cite{dhariwal:2020} for a deeper understanding.

\subsection{Quantization module}
Let us consider an input sequence $\mathbf{x} = x_1, \dots, x_T \in [-1,1]^T$  of length $T$, which represents a normalized waveform in the time domain. In order to be representative of an expressive portion of the audio sequence, $T$ should be large. However, due to the complexity of modern neural architectures, choosing a large enough value of $T$ is not always feasible. To reduce the dimensionality of the space one can leverage the VQ-VAE architecture \cite{oord2017} to map large continuous sequences in the time domain to smaller sequences in a discrete latent domain. A VQ-VAE is composed of three blocks: 
\begin{itemize}
    \item A convolutional encoder $E: [-1, 1]^T \to \mathbb{R}^{S\times D}$, with $S \ll T$, where $S$ is the  length of the latent sequence and $D$ denotes the number of channels;
    \item A bottleneck block $B=B_I \circ B_Q$ where  $B_Q: \mathbb{R}^{S\times D} \to \mathcal{C}^S \subseteq \mathbb{R}^{S \times D}$ is a vector quantizer, mapping the sequence of latent vectors $\mathbf{h}= \mathbf{h}_1, \dots, \mathbf{h}_S=E(\mathbf{x})$ into the sequence of nearest neighbors contained in a codebook $\mathcal{C} = \{\mathbf{e}_k\}^K_{k=1}$ of learned latent codes, and $B_I: \mathcal{C}^S \to [K]^S$ is an indexer mapping the codes $\mathbf{e}_{k_1}, \dots, \mathbf{e}_{k_S}$ into the associated codebook indices $z_1 = k_1, \dots, z_S = k_S$. Note that since $B_I$ is bijective, the codes $\mathbf{e}_k$ and their indices $k$ are semantically equivalent, but we shall use the term `codes' for the vectors in $\mathcal{C}$ and `latent indices' for the associated integers;
    \item A decoder $D:[K]^S \to [-1,1]^T$ mapping the discrete sequence back into the time domain.
\end{itemize}
The VQ-VAE is trained by minimizing the composite loss:
\begin{align}\label{eq:loss_vqvae}
\mathcal{L}_{\text{VQ-VAE}} &= \mathcal{L}_{\text{rec}} + \mathcal{L}_{\text{codebook}} + \beta \mathcal{L}_{\text{commit}}\,, \end{align} where:
\begin{align}
 \mathcal{L}_{\text{rec}} &= \frac{1}{T} \sum_{t}{\lVert x_t - D({z_t})\rVert^2_2} \\
 \mathcal{L}_{\text{codebook}} &= \frac{1}{S} \sum_{s}{\lVert \text{sg}[\mathbf{h}_s] - \mathbf{e}_{z_s}\rVert^2_2} \\
 \mathcal{L}_{\text{commit}} &= \frac{1}{S} \sum_{s}{\lVert \mathbf{h}_s - \text{sg}[\mathbf{e}_{z_s}]\rVert^2_2}\,, \end{align}
where $\text{sg}$ is the stop-gradient operator and $\beta$ is the commitment loss weight. The losses $\mathcal{L}_{\text{codebook}}$ and $\mathcal{L}_{\text{commit}}$ update the entries of the codebook $\mathcal{C}$ during the training procedure. In addition, we introduce a novel loss term $\mathcal{L}_{\text{lin}}$, described in Section $\ref{sec:likelihood}$, which imposes a precise algebraic structure on the latent space, facilitating the task of source separation.

\subsection{Latent autoregressive priors}
Once the VQ-VAE is trained, time domain data $\mathbf{x} \sim p^{\text{data}}$ can be mapped to latent sequences $\mathbf{z}$. Autoregressive priors $p(\mathbf{z})=p(z_1)p(z_2|z_1) \dots p(z_S|z_{S-1},\dots, z_1)$ can then be learned over the discrete domain.
In this work, the autoregressive models are based on a deep scalable Transformer architecture as in \cite{dhariwal:2020}.
In order to generate new time-domain examples, sequences of latent indices are sampled from $p(\mathbf{z})$ via ancestral sampling and then mapped back to the time domain via the decoder of the VQ-VAE. 
\section{Method}
\label{sec:method}

\begin{figure}[t!]
  \centering
    \def\svgwidth{\columnwidth}\scalebox{0.9}{
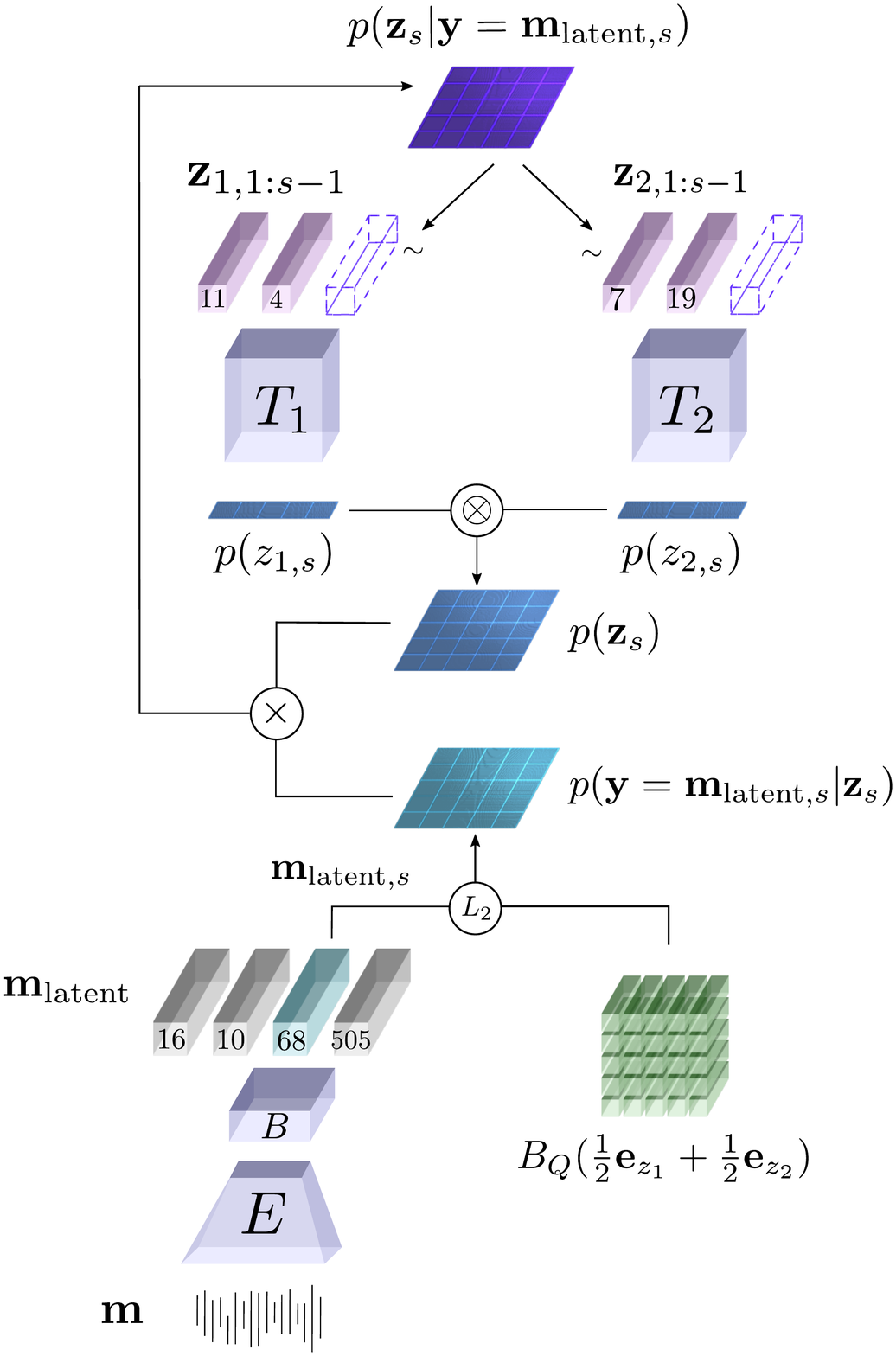}
\caption{In our method, two autoregressive priors $T_1$ and $T_2$ are trained on different instrument sources in the latent domain. At each step $s$ they provide the joint prior $p(\mathbf{z}_s)$. The prior is combined with a $\sigma$-isotropic Gaussian likelihood $p(y=\mathbf{m}_{\textnormal{latent}, s}|\mathbf{z}_s) = \mathcal{N}\left(\mathbf{m}_{\text{latent},s}\big| B_Q(\tfrac{1}{2}\mathbf{e}_{z_{1}} + \tfrac{1}{2}\mathbf{e}_{z_{2}}), \sigma^2 \mathbf{I}\right)$ in order to compute the posterior $p(\mathbf{z}_s\vert y = \mathbf{m}_{\textnormal{latent},s})$ from which new samples are drawn.}
\label{fig:method_1}

\end{figure}

\begin{figure}[t!]
  \centering
  \def\svgwidth{\columnwidth}\scalebox{1.0}{
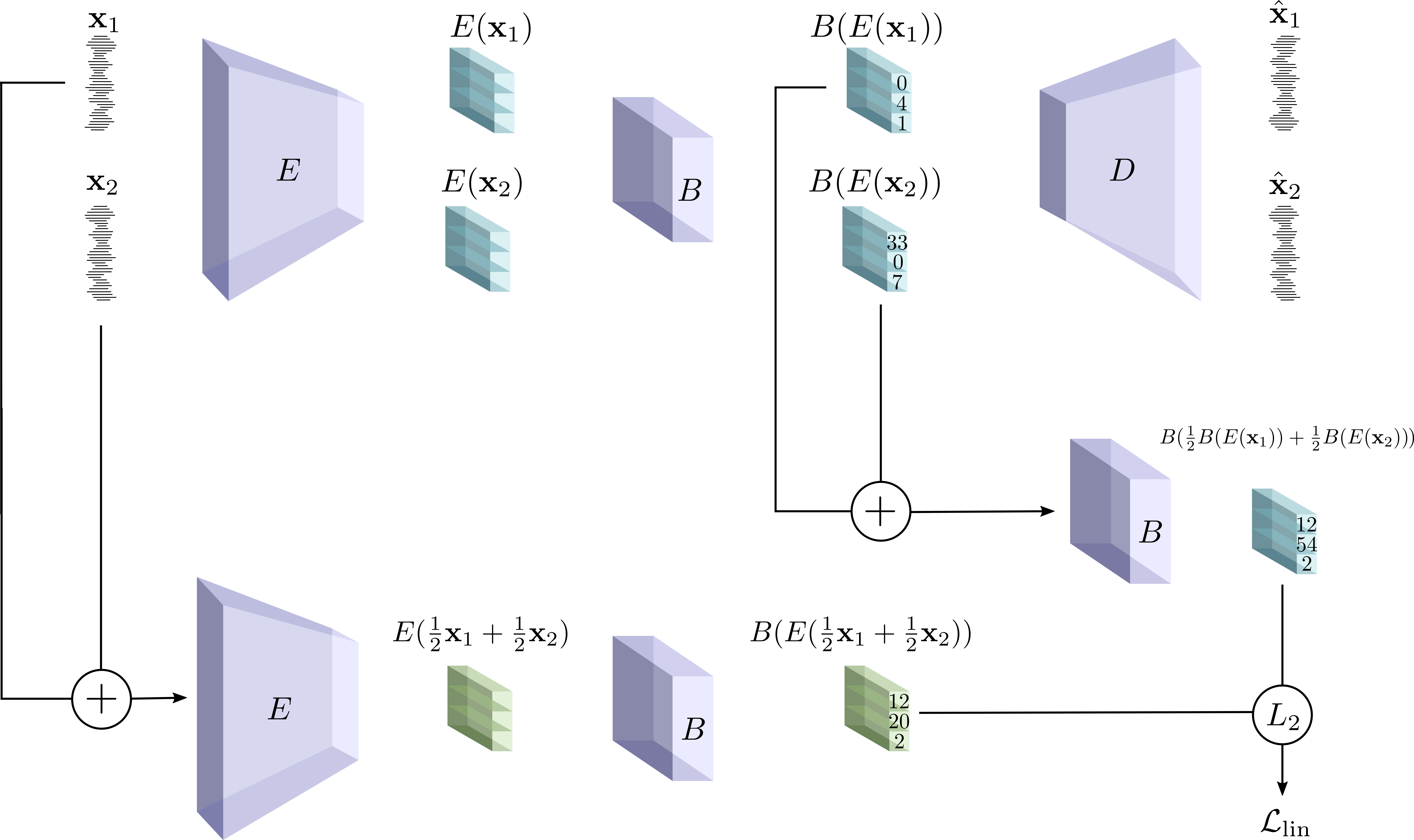}
\caption{
Training scheme of the LQ-VAE:
reconstructions $\hat{\mathbf{x}}_1$, $\hat{\mathbf{x}}_2$ are obtained from input pairs $\mathbf{x}_1, \mathbf{x}_2$ as in the VQ-VAE, leading to the loss $\mathcal{L}_\text{VQ-VAE}$ (Eq. $\eqref{eq:loss_vqvae}$). To this loss we add the post-quantization linearization loss $\mathcal{L}_{\text{lin}}$ (Eq. \eqref{eq:ll}), that is computed by matching time-domain sums with latent vector sums.}
\label{fig:method_2}
\vspace{-0.2cm}
\end{figure}

The proposed algorithm is composed of two parts. A first \emph{separation phase} in the latent domain, in which we sequentially sample from an exact posterior  on discrete indices.
A following \emph{rejection sampling procedure} based on a (scaled) global posterior conditioned on the separation results, which we use to sort the proposed solutions and select the most promising one.

\subsection{Latent Bayesian source separation}
     Our task is to separate a mixture signal $\mathbf{m} = \frac{1}{2} \mathbf{x}_{1} + \frac{1}{2}\mathbf{x}_{2}$ into $\mathbf{x}_1 \sim p_1^{\text{data}}$ and $\mathbf{x}_{2} \sim p_2^{\text{data}}$, where $p_1^{\text{data}}$ and $ p_2^{\text{data}}$ represent the distributions of each instrument class in the time domain. In a Bayesian framework, a candidate solution $\mathbf{x} = \mathbf{x}_1,\mathbf{x}_2$ is distributed according to the posterior $p(\mathbf{x}_1, \mathbf{x}_2\vert\mathbf{m}) \propto p_1^{\text{model}}(\mathbf{x}_1) p_2^{\text{model}}(\mathbf{x}_2) p(\mathbf{m} \vert \mathbf{x}_1, \mathbf{x}_2)$, where the priors $p_1^{\text{model}}$, $p_2^{\text{model}}$ are typically deep generative models and the likelihood $p(\mathbf{m}\vert \mathbf{x}_1, \mathbf{x}_2)$ is parameterized as $p(\mathbf{m}\vert \frac{1}{2}\mathbf{x}_1 + \frac{1}{2}\mathbf{x}_2)$.
     
     In this work, we follow the Bayesian approach but we work in the latent domain.
     After training the VQ-VAE on an {\em arbitrary} audio dataset (with samples lying also outside $p_1^{\text{data}}$ and $p_2^{\text{data}}$), we learn two latent autoregressive priors $p_1(\mathbf{z}_1)$ and $p_2(\mathbf{z}_2)$ over the two instrument classes. The priors do not require any correspondence between the sources, being trained in a completely unsupervised setting. We assume the two priors to be independent, i.e. $p(\mathbf{z}) = p(\mathbf{z}_1, \mathbf{z}_2) = p_1(\mathbf{z}_1)p_2(\mathbf{z}_2)$. Therefore, for each step $s \in [S]$, we can compute the posterior distribution 
          $p(z_{1,s}, z_{2,s} | \mathbf{z}_{1:s-1}, \mathbf{y}) \propto p_1(z_{1,s}|\mathbf{z}_{1,1:s-1})p_2(z_{2,s}|\mathbf{z}_{2,1:s-1})p(\mathbf{y}|z_{1,s},z_{2,s}, \mathbf{z}_{1:s-1})$.
          
          The random variable $\mathbf{y}=f(\mathbf{m})$ is a function of the mixture $\mathbf{m}$.
          One can choose to model $\mathbf{y}$ in multiple ways; a naive approach is to choose $f$ as the identity and set $\mathbf{y} = \mathbf{m}$, thus computing  the likelihood function directly in the time domain. This approach, however, requires the  decoding of at least $2K$ possible latent indices in order to locally compare the mixture $\mathbf{m}$ with the hypotheses $z_{1,s}$ and $z_{2, s}$. Note that this corresponds to a lower bound, given that the convolutional nature of the decoder requires a larger past context to produce meaningful results. Differently, we propose to define $\mathbf{y}$ in the latent domain, setting $\mathbf{y} = B_Q(E(\mathbf{m})) := \mathbf{m}_{\text{latent}}$. This approach is preferable since it does not require decoding the hypotheses at each step $s$, resulting in lower memory usage and computation time.
          Our method benefits from the choice of operating in the latent space, thanks to the relatively small size of the priors and the likelihood function domain (we choose $K = 2048$, as in \cite{dhariwal:2020}). 
          In addition, by exploiting the Transformer architecture, the prior distributions can be computed in parallel. For these reasons, evaluating and sampling from $p(z_{1,s}, z_{2,s} | \mathbf{z}_{1:s-1}, \mathbf{y})$  at each $s$ is computationally feasible and has $O(K^2)$ memory complexity. See Figure \ref{fig:method_1} for a visual description of the inference algorithm.

\subsection{Latent likelihood via LQ-VAE} \label{sec:likelihood}
 In this section we describe how we model the likelihood function and introduce the LQ-VAE model. Following \cite{jayaram2020} we chose a $\sigma$-isotropic Gaussian likelihood, setting:
 \begin{equation}
         \begin{array}{ll}
              \hspace{-0.8cm}p\left(\mathbf{m}_{\text{latent}}|z_{1,s},z_{2,s}, \mathbf{z}_{1:s-1}\right)  = \\ \hspace{0.7cm}= p\left(\mathbf{m}_{\text{latent},s}|z_{1},z_{2}\right)  \\ \hspace{0.7cm} = \label{eq:latent_likelihood}\mathcal{N}\left(\mathbf{m}_{\text{latent},s}\big| B_Q(\tfrac{1}{2}\mathbf{e}_{z_{1}} + \tfrac{1}{2}\mathbf{e}_{z_{2}}), \sigma^2 \mathbf{I}\right) \,.
          \end{array}
\end{equation}
 The hyper-parameter $\sigma$ balances the trade-off between the likelihood and the priors. Lower values promote the likelihood: the separated tracks combine perfectly with $\mathbf{m}$, but may not sound like the instrument of the class they belong to. Instead, higher values of $\sigma$ give importance to the priors: the separated tracks contain only sounds from the corresponding source distribution, but may not mix back to $\mathbf{m}$ (not resembling the sources).
The logarithm of the likelihood is: \begin{align}
               -\frac{1}{2\sigma^2}\left\lVert \mathbf{m}_{\text{latent},s} - B_Q\left(\tfrac{1}{2} \mathbf{e}_{z_{1}} + \tfrac{1}{2}\mathbf{e}_{z_{2}}\right) \right\rVert^2_2\,.
               \label{eq:sei}
           \end{align}
At each step $s$, we compare a variable term $\mathbf{m}_{\text{latent},s}$ with a constant matrix $B_Q\left(\tfrac{1}{2} \mathbf{e}_{z_{1}} + \tfrac{1}{2}\mathbf{e}_{z_{2}}\right)$ representing all possible (scaled) sums over all codes in $\mathcal{C}$. This term can be precomputed once and then reused during inference, saving additional computational resources.

%
We observed that performing separation with the likelihood in Eq. \eqref{eq:latent_likelihood} using a VQ-VAE trained with the loss in
Eq. \eqref{eq:loss_vqvae}, results in disturbed and noisy outcomes. Such behavior is expected because the standard VQ-VAE does not impose any algebraic structure on the discrete domain; therefore, summing codes as in Eq. \eqref{eq:latent_likelihood} does not lead to meaningful results. This problem can be lifted by enforcing a post-quantization linearization loss on the VQ-VAE: 
\begin{equation}
    \mathcal{L} = \mathcal{L}_{\text{VQ-VAE}} + \mathcal{L}_{\text{lin}} \,,
\end{equation}
where $\mathcal{L}_{\text{VQ-VAE}}$ is defined as in Eq. $\eqref{eq:loss_vqvae}$  and
\begin{align}
    \label{eq:ll}
    \mathcal{L}_\text{lin} &= \frac{1}{T} \sum_{t}{\left\lVert
    LQ_t - QL_t\right\rVert^2_2} \\ 
    \label{eq:ql}
    QL_t &=B_Q\left( \tfrac{1}{2}B_Q\left(E\left(\mathbf{x}_{1,t}\right)\right) + \tfrac{1}{2}B_Q\left(E\left(\mathbf{x}_{2,t}\right)\right)\right) \\ 
    \label{eq:lq}
    LQ_t &= B_Q\left(E\left(\tfrac{1}{2}\mathbf{x}_{1,t} + \tfrac{1}{2} \mathbf{x}_{2,t}\right)\right) \,.
\end{align}
Minimizing this loss pushes the quantized latent code representing a mixture of two arbitrary source signals ($LQ_t$ term) to be equal to the sum of the quantized latent codes, corresponding to the single sources ($QL_t$ term), therefore enforcing the discrete codes to behave in an approximately linear way.
We shall refer to the VQ-VAE trained as above, as a {\em Linearly Quantized Variational Autoencoder} (LQ-VAE). See Figure \ref{fig:method_2} for a visual illustration of the LQ-VAE training procedure.
\begin{table}[t!]
\centering
\setlength{\tabcolsep}{3pt}
\resizebox{\columnwidth}{!}{%
\begin{tabular}{lcccccl}
\hline
\multicolumn{1}{|l|}{\textbf{Method}}       & \multicolumn{1}{l}{\textbf{Drums}} & \multicolumn{1}{l|}{\textbf{Bass}}          & \multicolumn{1}{l}{\textbf{Drums}} & \multicolumn{1}{l|}{\textbf{Guitar}}        & \multicolumn{1}{l}{\textbf{Guitar}} & \multicolumn{1}{l|}{\textbf{Bass}}  \\ 
\hline
\multicolumn{1}{|l|}{Ours (best)}            & \textbf{5.83}          & \multicolumn{1}{c|}{\textbf{7.42}} & \textbf{8.33}             & \multicolumn{1}{c|}{3.80}          & 3.75             & \multicolumn{1}{c|}{\textbf{8.65}}  \\ 
\hline
\multicolumn{1}{|l|}{Ours (rej)} & 4.08                      & \multicolumn{1}{c|}{5.31}          & 6.93                      & \multicolumn{1}{c|}{2.48}          & 1.95                     & \multicolumn{1}{c|}{6.35}          \\ \hline
\multicolumn{1}{|l|}{Demucs$^{\dagger}$}            & 5.42                      & \multicolumn{1}{c|}{5.36}          &              5.80              &       5.36          &            \multicolumn{1}{|c}{6.42}       &
\multicolumn{1}{c|}{7.68}                                                     \\ \hline
\multicolumn{1}{|l|}{TasNet$^{\dagger}$}       & 5.51                      & \multicolumn{1}{c|}{5.43}                                  & \multicolumn{1}{c}{5.87}       &
\multicolumn{1}{c|}{\textbf{5.47}}& \multicolumn{1}{c}{\textbf{7.80}}       &
\multicolumn{1}{c|}{8.46}                                               \\ \hline

\multicolumn{1}{|l|}{rPCA\cite{rpca}}            & 0.60                      & \multicolumn{1}{c|}{1.05}          & 2.27                      & \multicolumn{1}{c|}{-0.42}          & 0.52                    & \multicolumn{1}{c|}{-1.12}          \\ \hline
\multicolumn{1}{|l|}{ICA\cite{ica}}           & -0.99                     & \multicolumn{1}{c|}{-1.53}         & \multicolumn{1}{c}{-0.53} & \multicolumn{1}{c|}{-3.23}         & -0.73                   & \multicolumn{1}{c|}{-2.79}        \\ \hline
\multicolumn{1}{|l|}{HPSS \cite{hpss}}            & -0.56                     & \multicolumn{1}{c|}{-0.33}          & 0.31                     & \multicolumn{1}{c|}{-2.72}          & 0.15                     & \multicolumn{1}{c|}{-0.38}          \\ \hline
\multicolumn{1}{|l|}{REPET\cite{repet}}          & 0.53                      & \multicolumn{1}{c|}{1.54}           & 2.91                      & \multicolumn{1}{c|}{0.11} & 0.40                    & \multicolumn{1}{c|}{-1.09}           \\ \hline
\multicolumn{1}{|l|}{FT2D \cite{ft2d}}            & 0.59                      & \multicolumn{1}{c|}{1.31}          & 2.63                      & \multicolumn{1}{c|}{-0.15}          & 0.65                    & \multicolumn{1}{c|}{-1.02}         \\ \hline
\end{tabular}
}
\vspace{0.1cm}
\caption{SDR scores evaluated on Slakh2100 test set. All methods are unsupervised except those marked with ${\dagger}$. The \textnormal{rej} attribute indicates that the solutions were obtained by the rejection sampling procedure with $\alpha = 0$. The scores are computed according to the implementation in \cite{SiSEC18}}.
\label{table:sdr}
\end{table}

\begin{table}[t!]
\setlength{\tabcolsep}{3pt}
\centering
\resizebox{\columnwidth}{!}{%
\begin{tabular}{lcccccl}
\hline
\multicolumn{1}{|l|}{\textbf{Rejection $\alpha$}}       & \multicolumn{1}{l}{\textbf{Drums}} & \multicolumn{1}{l|}{\textbf{Bass}}          & \multicolumn{1}{l}{\textbf{Drums}} & \multicolumn{1}{l|}{\textbf{Guitar}}        & \multicolumn{1}{l}{\textbf{Guitar}} & \multicolumn{1}{l|}{\textbf{Bass}}  \\

\hline
\multicolumn{1}{|l|}{0}            & \textbf{4.08}          & \multicolumn{1}{c|}{\textbf{5.31}} & \textbf{6.93}             & \multicolumn{1}{c|}{\textbf{2.48}}          & \textbf{1.95}             & \multicolumn{1}{c|}{\textbf{6.35}}  \\ 

\hline
\multicolumn{1}{|l|}{0.5} &3.61                      & \multicolumn{1}{c|}{4.78}          & 6.69                      & \multicolumn{1}{c|}{2.17}          & 1.68                     & \multicolumn{1}{c|}{6.00}          \\ \hline

\multicolumn{1}{|l|}{1}            & 2.94                      & \multicolumn{1}{c|}{4.03}          &              6.44              &       1.95         &            \multicolumn{1}{|c}{1.15}       &
\multicolumn{1}{c|}{5.35}                                                      \\ \hline
\end{tabular}}
\caption{Ablation study for rejection parameter $\alpha$.} 
\label{table:rejection_ablation}
\end{table}
\begin{table}[t!]
\centering
\begin{tabular}{lccl}
\hline
\multicolumn{1}{|l|}{\textbf{Method}}       &  \multicolumn{1}{l}{\textbf{Drums}} & \multicolumn{1}{l|}{\textbf{Piano}}         \\ 
\hline
\multicolumn{1}{|l|}{Ours}            &  \textbf{0.68}            & \multicolumn{1}{c|}{\textbf{3.66}} \\ 
\hline
\multicolumn{1}{|l|}{Ours (rejection $\alpha=0$)} & 0.08                     & \multicolumn{1}{c|}{2.75}          \\ \hline
\multicolumn{1}{|l|}{GAN [20]}             & -3.16                     & \multicolumn{1}{c|}{-2.26}         \\ \hline
\end{tabular}
\caption{SDR table evaluated on the test set of \cite{narayanaswamy2020unsupervised}.}
\label{table:gan}
\vspace{-0.9cm}
\end{table}

\subsection{Rejection sampling}

Given the low memory requirements of our method, at inference time we can sample in parallel multiple solutions $\{\mathbf{z}^{(b)}\}_{b=1}^{B}$ in the same batch. Autoregressive models tend to accumulate errors over the course of ancestral sampling, therefore the quality of the solutions varies across the batch. 
In order to select a solution, we look at the posterior $p_{\text{rej}}(\mathbf{z}\vert\mathbf{m}) \propto p_{\text{rej},1}(\mathbf{z}_1)p_{\text{rej},2}(\mathbf{z}_2)p_{\text{rej}}(\mathbf{m}\vert\mathbf{z})$, conditioned by the sampling event. We obtain the priors $p_{\text{rej},1}$ and $p_{\text{rej},2}$ by normalizing $p_1$ and $p_2$ over the batch (computed by integrating over $s$ during the inference). For numerical stability, we scale their logits by the length of the latent sequences $S$. The likelihood function $p_{\text{rej}}(\mathbf{z}\vert\mathbf{m}) = \mathcal{N}\left(\mathbf{m}\big\vert \tfrac{1}{2}D(\mathbf{z}_1) + \tfrac{1}{2}D(\mathbf{z}_2), \sigma_{\text{rej}}^2\mathbf{I}\right)$ is computed directly in the time domain, with the decoding pass being executed only once at the end of the sampling procedure. The hyper-parameter $\sigma_{\text{rej}}$ plays a similar role to the $\sigma$ used in Eq. \eqref{eq:latent_likelihood}. We can balance the likelihood and the priors by setting:
\begin{equation*}
    \mathbb{E}_{b}\hspace{-0.1cm}\left[\log p_{\text{rej}}(\mathbf{z}^{(b)})\right] \hspace{-0.1cm}=\hspace{-0.1cm} -\frac{1}{2\sigma_{\text{rej}}^2} \mathbb{E}_{b}\hspace{-0.1cm}\left[
               \left\lVert \mathbf{m} \hspace{-0.08cm}- \hspace{-0.08cm}
               \tfrac{1}{2}(D(\mathbf{z}^{(b)}_1) + D(\mathbf{z}^{(b)}_2))
               \right\rVert^2_2
           \right]
\end{equation*} and solving for $\sigma_\text{rej}$. Albeit natural, this framework does not lead to the best selection. We performed an ablation study by weighting the contribution of the global likelihood with a scalar $\alpha \in [0, 1]$ (using ${{\sigma'}}_{\text{rej}}^2 = \alpha \sigma^2_{\text{rej}} $) and the best empirical results are obtained when the global likelihood is not taken into account ($\alpha = 0)$, see Table \ref{table:rejection_ablation}.  We call this selection criterion \textit{prior-based rejection sampling}.

\section{Results}
\label{sec:experiments}

We validate our approach on \textit{Slakh2100}  \cite{manilow2019}: 
a large musical source dataset containing mixed tracks separated into $34$ instrument categories. We select tracks from the classes `drum', `bass' and `guitar' coming from the training and test splits, sub-sampled at a frequency of $22$kHz. We train the convolutional LQ-VAE over mixtures obtained by randomly mixing sources from the individual tracks of the training set. The LQ-VAE has a downsampling factor of $\frac{T}{S} = 64$ and uses a dictionary of $K = 2048$ latent codes. After training the LQ-VAE, we train two autoregressive models, one per source, on latent codes extracted from $\sim 1200$ tracks each. In all our separation experiments we fixed $\sigma = 0.1$ in Eq. \eqref{eq:sei}.
%
%
In Table \ref{table:sdr} we compare our method with two state-of-the-art supervised approaches and different non-learning based unsupervised methods. To this end, we iterate on the test split of \cite{manilow2019} made up of about $150$ different songs, and for each we extract $450$ random chunks each of $3$ seconds. 

In order to strengthen our empirical evaluation, we show in Table \ref{table:gan} results of our model applied to a different validation data set in order to perform a comparison with the GAN model of \cite{narayanaswamy2020unsupervised}. We evaluate both methods over the test dataset
proposed in \cite{narayanaswamy2020unsupervised}, consisting of 1000 mixtures of 1 second each. Each mixture combines a drum sample with a piano track randomly, thus independence
in the test data is assumed, resulting in a more artificial setting with respect
to the one present in Slakh2100. For \cite{narayanaswamy2020unsupervised} we use the pre-trained model
given by the authors while for our method we use the “drums” and “piano” priors trained on Slakh2100 thus showing the cross-dataset generalization
capability of our model.
%

All our experiments are performed on a Nvidia RTX 3080 GPU with $16$ GB of VRAM. With this GPU our method can sample a batch of $200$ candidate solutions ($100$ for each instrument) simultaneously. The code to reproduce our experiments is available at \url{https://github.com/michelemancusi/LQVAE-separation}.
Interestingly, even if solutions selected by the rejection sampling algorithm have slightly lower metrics than supervised approaches, by individually selecting the best solution for each instrument we achieve performance in line with the state of the art (especially on `bass` and `drum` stems). This testifies the quality of our separation. 
%
Remarkably, our method employs $3$ minutes on average for sampling a track of $3$ seconds,
compared to the more than $100$ minutes of \cite{jayaram2021}.

\section{Conclusions}
\label{sec:conclusion}
In this work, we introduced a simple algorithm to perform exact Bayesian inference in the discrete latent domain.
Our method allows to achieve good separation results while being much faster than other likelihood-based unsupervised approaches.

The main bottleneck of our method lies in the rejection sampling strategy. Future work will attempt to improve this aspect by investigating the design of more accurate learning-based rejection samplers. Other benefits could come from the adoption of multi-level VQ-VAEs \cite{dhariwal:2020} or by leveraging deeper autoregressive priors.


\bibliographystyle{IEEEtran}
\bibliography{bibliography}

\end{document}